\documentclass{sig-alternate-05-2015}
\usepackage{color}
\usepackage{subcaption}
\usepackage{algorithm2e}
\usepackage{multirow}
\usepackage{float}
\usepackage{graphicx}
\graphicspath{ {images/} }

\begin{document}

\title{Which techniques does your application use?:\\ An information extraction framework for scientific articles}
\author{Soham Dan, Sanyam Agarwal, Mayank Singh, Pawan Goyal and Animesh Mukherjee\\
       \affaddr{Department of Computer Science and Engineering}\\
       \affaddr{Indian Institute of Technology, Kharagpur, WB, India}\\
       \email{\{sohamd,sanyama\}@cse.iitkgp.ernet.in}\\
	\email{\{mayank.singh,pawang,animeshm\}@cse.iitkgp.ernet.in}
}
\date{16 July 2015}
\maketitle
\begin{abstract}
Every field of research consists of multiple {\em application areas} with various {\em techniques} routinely used to solve problems in these wide range of application areas. With the exponential growth in research volumes, it has become difficult to keep track of the ever-growing number of application areas as well as the corresponding problem solving techniques. 
In this paper, we consider the computational linguistics domain and present a novel information extraction system that automatically constructs a pool of all application areas in this domain and appropriately links them with corresponding problem solving techniques. Further, we categorize individual research articles based on their application area and the techniques proposed/used in the article. $k$-gram based discounting method along with handwritten rules and bootstrapped pattern learning is employed to extract application areas. Subsequently, a language modelling approach is proposed to characterize 
each article based on its application area. Similarly, regular expressions and high-scoring noun phrases are used for the extraction of the problem solving techniques. We propose a greedy approach to characterize each article based on the techniques. Towards the end, we present a table representing the most frequent techniques adopted for a particular application area. Finally, we
propose three use cases presenting an extensive temporal analysis of the usage of techniques and application areas. 
\end{abstract}
\newline
\newline


\newline


\section{Introduction}
It is not uncommon for researchers to envisage an information extraction system for scientific articles that can answer queries like, (i)\textit{what are all the techniques and tools
used in machine translation?}, (ii)\textit{Which are the subareas of computational linguistics, where Malt
parser is frequently used?} etc. However, the meta-information necessary for constructing such a system is rarely available. Each research domain consists of multiple application areas which are typically associated with various techniques 
used to solve problems in these areas. Knowledge of these typical techniques adopted for an application area is of crucial importance to a researcher 
focusing on that application area. For instance, two commonly used techniques in \textbf{information extraction} are \textit{conditional random fields} and 
\textit{hidden markov models}. 
Answering queries such as to obtain list of significant techniques given an application area or to obtain list of all application areas given a technique
is extremely valuable to every researcher, but particularly for those who are venturing into a new area. An automated information extraction system would save humongous manual efforts to come up with such a list for each research field. Moreover, the lists are dynamic in nature, which further presents challenges in generation and automation. New techniques are added to the area with time and changing needs. Thus, it is of interest to the researcher to know the significant techniques for an area within a time frame. Thus, the temporal aspect is also of particular interest and raises diverse research questions - for example, how have the techniques for "part-of-speech tagging" varied over time, or, what are the most important areas of computational linguistics that have been addressed in the last 5 years? 

In this work, 
%due to limitations in the availability of full-text data and computational constraints, 
we demonstrate the system for the domain of computational linguistics because of the availability of full-text research articles. However, the proposed schema can be easily generalized to any other field of research. Next, we define two common keywords used
in the current paper:

\noindent\textbf{Area}: Area represents an application area of computational linguistics domain. Common application areas include `machine translation', `dependency parsing', `part-of-speech tagging', `information extraction', `question answering', etc.

\noindent\textbf{Technique}:  Technique either represents a tool or a task used in an area. Common examples include `Bleu score', `Rouge score', `Charniak parser', `TnT tagger', `Malt parser', `MST parser', etc. Note that \textit{technique} of one paper can potentially be an area of another paper. For example, in ``Training Nondeficient Variants of IBM-3 and IBM-4 for Word Alignment''~\cite{schoenemann2013training}, `word alignment' is an area but in ``Using Word-Dependent Transition Models in HMM-Based Word Alignment for Statistical Machine Translation''~\cite{he2007using}, `word alignment' is a technique for machine translation.

In section~\ref{method}, we describe an algorithmic approach to construct a mapping between a list of areas and a list of techniques which is further used to construct a summary table of the meta-data.
The approach is organized into five important steps: 
\begin{enumerate}
\item We first create a ranked list of application areas;
\item Given a research paper, we automatically assign it to one of the areas based on its title and abstract; 
\item Next, we create a ranked list of techniques;
\item Given a paper, we automatically extract all the techniques it introduces; 
\item Finally, we automatically construct a mapping from the list of areas to the list of techniques. 
\end{enumerate}

%As an additioanl analysis, we perform a temporal analysis on the data. Dividing the time-line into windows we show typical examples of techniques which have been popular or fallen out of use for a given application area. 
We achieve significantly high performance in each of the above steps. The precision of the first step is \textbf{84\%} (for top 25) and recall is \textbf{87\%}. For the second step, the accuracy is \textbf{73.3\%}. The third step results in a precision and recall of \textbf{80\%} (for top 25) and \textbf{81\%} respectively. In the fourth step, our system achieves an accuracy of \textbf{60\%}. Finally, for the fifth step we present a number of relevant examples from the entire mapping. 
As a use case for this study, we analyze temporal characteristics of the techniques (Section \ref{use-case}). In an area, over the time, certain 
techniques become quite popular (as determined by the number of times that technique is cited by any paper) while certain techniques fall out of favor. Advent of social media is an example of this phenomenon where, due to different writing styles (short text length, use of out-of-vocabulary words, character flooding etc.), the need of new techniques is particularly essential. We further explore other aspects under temporal analysis -- how the popular (as determined by the number of papers published in that area) areas in computational linguistics vary with time. We also investigate the temporal variation of the popular areas for specific conferences, namely, \textsc{acl} and \textsc{coling}.

\iffalse{}
With the introduction of social media and subsequent large-scale research on social media data, there was a need for new
techniques ( example : tools for part-of-speech tagging) for a number of areas  of interest for social media analysis (example : part-of-speech tagging). 
Standard POS taggers did not work on social media data as well as it did on typical, clean text data due to short text, use of out-of-vocabulary characters,
character flooding and other text modifications. Thus, we expect a shift in techniques for part-of speech tagging and certain other areas, around the time 
social media research became prominent.  \newline

Contribution : In this paper, we have depicted the stages that are necessary to construct a mapping from areas to techniques. Broadly, we introduce novel
methods for extraction of the pool of areas and techniques for the computational linguistics domain and for the categorization of individual papers by the 
area they belong to and the techniques they introduce or improve upon. Lastly, we show that a temporal analysis on this data can yield many interesting 
results. Our work is novel in its attempt to answer these significant research questions .
\fi

\section{Related Work}
Extracting application area and techniques is primarily an information extraction task. 
Information extraction (IE) from scientific articles combines approaches from natural language processing and data mining 
and has generated substantial research interest in recent times. In particular, there has been burgeoning research interest 
in the domain of biomedical documents. Shah et al.~\cite{Shah2003} extracted keywords from full text of biomedical articles and 
claim that there exist a heterogeneity in the keywords from different sections. Muller et al.~\cite{muller2008textpresso} have developed the \textit{Textpresso} framework, 
that leverage ontologies for information retrieval and extraction. In a similar work, Fukuda et al.~\cite{fukuda1998toward} proposed an IE system for protein name extraction. There has been significant work in information extraction in the area of protein structure analysis. 
Gaizauskas et al.~\cite{gaizauskas2003protein} proposed \textit{PASTA}, an IE system developed and evaluated for the protein structure domain. 
Friedman et al.~\cite{friedman2001genies} have developed a similar system to extract structure information about cellular pathways using a knowledge model. 
Biological information extraction has seen extensive work covering diverse aspects with large number of survey papers. Cohen et 
al.'s~\cite{cohen2005survey} survey on biomedical text mining, Krallinger et al.'s~\cite{krallinger2008linking} survey on 
information extraction and applications for biology and Wimalasuriya et al.~\cite{wimalasuriya2010ontology} on ontology based 
information extraction are examples of some of the popular surveys on IE for biomedical domain.

Information extraction in other domains have also received an equally strong attention from researchers. Hyponym relations have been 
extracted automatically in the celebrated work by Hearst et al.~\cite{Hearst:1992:AAH:992133.992154}. Caraballo et al.~\cite{caraballo1999automatic}
have extended previous work of automatically building semantic lexicons to automatic construction of a hierarchy of nouns and their hypernyms. 
Teufel~\cite{teufel2000argumentative} proposed information management and 
information foraging for researchers and introduced a new document analysis technique called argumentative zoning which is useful for generating user-tailored 
and task-tailored summaries. Kim et al.~\cite{kim2010semeval} and Lopez et al.~\cite{lopez2010humb} are two popular works in automatic keyphrase extraction from 
scientific articles. Quazvinian et al.~\cite{qazvinian2008scientific} have explored summarization of scientific papers using citation summary networks and citation summarization through keyphrase extraction~\cite{qazvinian2010citation}.

Jones~\cite{jones2005learning} introduced an approach for entity extraction from labeled and unlabeled text. They proposed algorithms that alternately 
look at noun phrases and their local contexts to recognize members of a semantic class in context. A relatively recent work by Gupta et
al.~\cite{gupta2014spied} developed a pattern learning system with bootstrapped entity extraction.In ~\cite{gupta2011analyzing}, they investigated the dynamics of a 
research community by extracting key aspects from scientific papers and showed how extracting key information help in analyzing the 
influence of one community on another. %%%%%Banko et al.~\cite{mintz2009distant} introduced the concept of open information extraction from the web, that performs better than the state-of-the-art Web IE systems. 
Jin at al.~\cite{jin-EtAl:2013:EMNLP} proposed a supervised
sequence labeling system that identifies scientific terms and their accompanying definition. 

In our work, we have developed and tested an information extraction system that can extract the area and techniques from scientific papers and build a repertoire of such areas and techniques from the corpus. Further, we have constructed a mapping from each area to the list of techniques used for that area. As a use case for our developed system, we do a thorough temporal analysis for the variation of techniques for an area with time, temporal variation of popular areas for the entire AAN dataset and temporal variation of popular areas for top conferences in computational linguistics-namely, \textsc{acl} and \textsc{coling}.

\section {Dataset}
We use AAN (ACL Anthology Network)~\cite{Radev} dataset which consists of 21,213 full text papers from the domain of computational linguistics and natural language processing. The dataset consists of papers between the years 1965 -- 2013 from 342 ACL venues. We further pre-process the full text articles to remove OCR errors. Dataset and code is available online at http://tinyurl.com/hhrpfge.

\section{Methodology}
\label{method}
In this section, we describe a method to construct a mapping between a list of areas and a list of techniques.
As we already pointed out in the introduction, the mapping task is organized into five steps: (1) creation of a ranked list of areas, (2) categorizing papers on the basis of areas, (3) creation of a ranked list of techniques, (4) categorizing papers on the basis of techniques, and (5) the final mapping between the list of areas to the list of techniques.

The current work has two-fold contribution. First, we develop a system which automatically extracts area and techniques used in a paper. Second, we create a mapping from area to corresponding popular techniques. Together, they can be thought of as structured metadata for individual scientific articles as well as a research field. 

Next, we briefly describe five phases in further details:

\subsection{Building the repertoire of areas}

We employ paper title information to extract areas. AAN provides paper title information as a metadata. We use hand-written rules to extract phrases which are likely to contain the area names. We observe that some functional keywords, such as, ``for'', ``via'',``using'' and ``with'' act as delimiters for such phrases. 
For example, paper title, ``Moses: Open source toolkit for statistical machine translation''~\cite{koehn2007moses} represents an instance of the form \textit{X for Y}, where Y is 
the application area. We also observe that the phrase succeeding ``for'' or preceding ``using'' or both (e.g., in ``Decision procedures for dependency parsing using graded constraints'' \cite{menzel1998decision}) are likely to contain the name of an application area. 

\noindent{\bf Seed set creation:} We create a seed set of the above functional keywords and use bootstrapped pattern learning to gather more such words along with areas.
For example, given a word `\textit{via}' in the seed set and paper title ``Non-Monotonic Sentence Alignment via Semisupervised Learning''~\cite{quan2013non}, we extract the
leading phrase ($k$-gram) before `\textit{via}'. Further, we extend the seed set by extracting more functional keywords. For example, given a paper title ``Improving English Russian sentence alignment through POS tagging and Damerau Levenshtein distance''~\cite{kutuzov2013improving} and the above extracted phrase ($k$-gram) \textit{alignment}, we enrich the seed set by the functional keyword `\textit{through}'. We had initially started with seven functional keywords and by bootstrapped pattern learning, augmented this to a final set of 11 functional keywords.

\noindent{\bf Ranking of the extracted phrases:} The previous step helps in extracting all the possible candidate phrases, which could potentially be {\sl area} names. This set contains a lot of noisy phrases such as, ``machine translation system combination and evaluation''. %(\textit{W11-2137}). 
Here, ``machine translation'' must be extracted from the surrounding noisy words. Next, we use empirical ranking algorithms to extract the exact area names from this collection.% of phrases using empirical ranking algorithms.
We observe that empirical ranking algorithms produce good results in extraction of the exact area names from long phrases. We employ three ranking schemes, described below:
\begin{itemize}
\item \textbf{Scheme 1:} In this scheme, we rank according to individual $k$-gram scores. The score of a given $k$-gram ($K$) is calculated as:
 \begin{equation}
  Score_K = \frac{count_K}{ \sum_j{count_j}}
 \end{equation}
where, $count_K$ represents occurrence count of the $K^\textrm{th}$ $k$-gram and the denominator represents total count of all the $k$-grams. 
\item \textbf{Scheme 2:} In this scheme the scoring method is very similar to previous scheme. However, an additional  heuristic employed is that if the score of a $k$-gram is greater than both of its border $(k-1)$ 
grams then the border $(k-1)$ grams are left out.  The intuition behind this is as follows: the trigram ``word sense disambiguation" will have a higher score than its border bigrams, ``word sense'' and ``sense disambiguation'', causing these bigrams to be left out.
\item \textbf{Scheme 3:} This is an improvement on the previous method in that different threshold scores are chosen for bigrams, trigrams, 4-grams and 5-grams below 
which they are eliminated. All higher order $k$-grams are also eliminated.  Unigrams were also not selected based on a pilot experiment with only unigrams which indicated that application areas are not usually unigrams. The thresholds were selected manually by observing the individual lists for each value of $k = 2, 3, 4$ and $5$.
In the result section, we shall compare the precision of each of these methods and we have finally adopted Scheme 3 since it gives the best results.
We present 24 of the top 30 areas judged accurate by domain experts:\end{itemize}
\fbox{%
    \parbox{0.45\textwidth}{%
        machine translation, natural language processing, word sense disambiguation, speech recognition, question answering, dependency parsing, information extraction, chinese word segmentation, semantic role labeling, information retrieval, entity recognition, word alignment, conditional random fields, maximum entropy, coreference resolution, machine learning, dialogue systems, textual entailment, natural language understanding, active learning, part-of-speech tagging, relation extraction, sentiment analysis, sense induction
    }%
}

%For this step, we measure precision in the next section. 

\subsection{Categorizing papers on the basis of application area}
\begin{center}
\begin{figure}[h]
\includegraphics[width=0.5\textwidth, height=5.5cm]{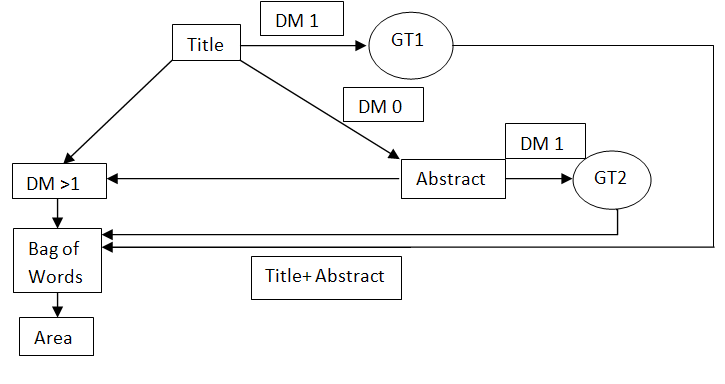}
\caption{Flowchart for the algorithm to extract areas from individual papers. DM $n$ stands for direct match with $n$ areas. }
\label{workflow}
\end{figure}
\end{center}

From the first step, we obtain a pool of application areas and we have to assign individual papers to one of these areas. Individual papers from the entire corpus are categorized to their corresponding  areas on the basis of two strategies -- direct match and relevance as per the language models, defined for various areas (see Figure~\ref{workflow}).  
\noindent{\bf Direct match:} In the direct match approach, we search for an explicit string match between the title or abstract and one of the areas. One non-trivial task here is identifying the location and the text in the abstract of a paper. We achieve this by using handwritten rules $[digit]?["abstract"]$ upto $[digit]?["introduction"]$ (where the "?" denotes that the section number and is optional), to match the portion of the paper between the abstract and the introduction. Also, we have converted the text to lower case during pre-processing. Subsequently, using the set of areas obtained from the previous step, we search if there is a direct match between any of these and the title. If the given title contains only one area name from the previous list, then the paper is assigned to that area. In case we do not find a match in the title, we check for a direct match with the abstract of the paper. If the abstract contains only one such matching area then the paper is categorized to that area. On the other hand, if the title or the abstract contains more than one direct match with the area set then we further use the language modeling approach to classify that paper. 

 \noindent{\bf Language modeling for classification of area of individual paper:} In this approach, we create a language model for each area, and classify a document into one of these areas based on the probability that the title and abstract of the document is generated from the language models of various areas.% Thus, he documents are equivalent to the areas in our context and the title and abstract of the paper to be categorized is our query. 
To create a language model for each area, we select the papers which could be classified on the basis of a single direct match. The titles and abstracts of all the papers belonging to one area are taken together to construct the language model of that area %%%A bag of words is maintained for each document consisting of all the words contained in the title and abstract of these papers. The bag of words is maintained as term and its corresponding count for ease of access.
with the Jenilek-Mercer (JM) smoothing.
%based language model and calculate the score corresponding to each area. 
A document not categorized using direct match is treated as a query, consisting of the words in its title and abstract. The Porter stemmer is used on the bag-of-words for each area and the query. As validated by ~\cite{zhai2004study}, the parameter $\lambda$ for JM smoothing should be high for long queries, which is true in our case. After experimenting on a small set of sample of papers, we fixed $\lambda$ to 0.7. %Since, $log(P(q|d))$ is given as the sum of the term weights = $log( 1+ \frac{(1-\lambda)P_{ml}(q_{i}|d)}{\lambda P(qi|C)})$, for each $q_{i}$. $P(q|d)$ is given as the product of the arguments of the logarithms of the term weights for each $q_{i}$. 
Now, in our context, the prior probability $P(d)$ for an application area, is proportional to the number of papers which were assigned to that area by a single direct match of either the title or the abstract. Hence, given a query paper $q$, the area which scores the highest $P(d|q)=P(q|d)P(d)$ for the query ($q$) is assigned as the area for the given paper.%%% Note that as per the definition of this smoothing technique, $P(q|d)$ is given by the product of the arguments of the logarithms of the term weights for each query.

\subsection{Extracting the repertoire of techniques}
The central idea here is based on the notion of {\em method} papers:  these are papers that introduce a novel technique or provide a toolkit in an area of computational linguistics. For instance, the paper introducing the Stanford CoreNLP toolkit is one such relevant example. The novel technique introduced or improved upon is a reason why method papers are cited. These papers are thus characterized by two features -- one, they are expected to have been cited a number of times which is above some threshold  (say $k_1$) thus indicating that the technique introduced or improved upon is frequently used and second, the fraction of times they obtain their citations in the ``methodology'' section of other papers is above some threshold (atleast $k_2$\%), %of their total number of citations, 
thereby, indicating that they are primarily ``method papers". We have set these thresholds by observing a large subset of papers of the ACL corpus. The values we chose to identify method papers are $k_1 = 15$ and $k_2 = 50$\%. 

\noindent{Identifying method papers:} one non-trivial task in coming up with the list of method papers is to identify the method section from the text of a research paper. For this, we searched the text with the pattern [digit][S] where S is a string containing some keywords like ``methodology" ,``method", ``approach" or syntactic variations of these. The digit indicates the section number. This step help us identify as to which of the citations occur in the method section. Now, once the ``method papers" are identified using the two thresholds above, we need to associate them with the techniques they are used for. For example, the Stanford CoreNLP toolkit paper will be cited whenever the citing paper uses the former's part-of-speech tagging module, tokenizer module or the named-entity recognition module, although these are quite different techniques. Hence we need to associate or extract from the Stanford CoreNLP toolkit paper the techniques, part-of speech tagging, tokenization, named entity recognition, among other techniques.  %Here we again need to define a term- Citation context : This is the sentence in which the citation is made. %

Here we make use of the fact that when the citing paper applies a technique from the cited paper, it cites that paper and also mentions the name of the technique in the {\em citation context} (i.e., the sentence where the citation is made). Our objective is to extract all the techniques a method paper is used for, from the citation context(s). In other words, given all the citation contexts for a particular ``method paper", we wish to identify all the specific techniques for which that method paper is used. We now describe the algorithm in detail.

\noindent{\bf Step 1:} As a first step, we need to identify the citations in the text of the paper. One example of a citation is: ``(String et al., year)''. There are numerous forms in which the citations could be written. Sometimes the year is shortened to two digits or if there are only two authors, their names are separated by ``and" and so on. Further, we had to take into account the possible errors in the text like missing parenthesis. We hand-crafted a set of regular expressions to match the citation format. We tested these on a small sample of 30 papers and augmented the set. The final set was used for extracting the citation contexts. 

\noindent{\bf Step 2:} For every method paper in the corpora, we extract all its citation contexts when the citation is made in the method section. %Thus, if paper Y cites method paper X in the method section, only then we add the citation context from paper Y. We also generate an individual file for each method paper containing only the citation contexts for that paper. Thus, we have all the method citation contexts for every paper as well as a single large file having all the method citation contexts across the whole corpora.
One observation we make is that the techniques are almost exclusively noun phrases in the citation context. We build a global vector of noun phrases across all citation contexts for all the method papers. The $i$$^\textrm{th}$ component of the vector is the raw count of the $i$$^\textrm{th}$ noun phrase, ordered lexicographically, over the method citations of the entire corpora. We consider this global vector as the ranked list of all the techniques used in the computational linguistics domain. Some of the top ranking noun phrases are depicted below.

\noindent\fbox{%
    \parbox{0.45\textwidth}{%
        penn treebank, stanford parser, rate training, berkeley parser, machine translation, statistical machine translation, charniak parser, moses toolkit, word sense disambiguation, maximum entropy, ibm model, bleu score, perceptron algorithm, word alignment, stanford pos tagger, collins parser, natural language processing, bleu metric, coreference resolution, moses decoder, giza++ toolkit, brill tagger, tnt tagger,anaphora resolution, mst parser, ccg parser, malt parser, minimum error rate training
    }%
}
\subsection{Extracting techniques for an individual\\ method paper}
For  each method paper, X, we extracted all the noun phrases that have been used in its citation context(s). We built a similar vector of these noun phrases where the $i$$^\textrm{th}$ component of the vector is the raw count of that noun phrase drawn from the global vector introduced in the previous section. If a particular noun phrase from the global vector is missing in the citation contexts for X, its weight is set to zero. Once this vector is created for method paper X, we simply take a dot product between this local vector of X and the global vector to get a ranked list of possible techniques for X. Now, we simply read off the top $K$ techniques on this rank list as the techniques the method paper X is used for. 

However, note that since we have taken a dot product between the local and global noun phrase vectors, those noun phrases which occur frequently across almost all papers might wrongly turn up as a technique if it occurs even once in the local vector. Noun phrases such as ``citation", ``previous work" and ``recent work" are some typical examples of such occurrences that we found in our experiment. Thus, we maintain a stop-phrase list of noun phrases that commonly occur in the citation context but are not techniques. Further, we ignore the technique which exactly matches with, contains or is contained in the area since this is a redundant case. For example, for the application area "Statistical Machine Translation", we ignore the technique "Machine Translation". After this stage we were able to extract a list of top $K$ techniques for each method paper.

\subsection{Mapping between area and the list of \\techniques}
Construction of the mapping table is the final and the most important step of our problem. For building this mapping we follow a simple method of count updating.
\begin{algorithm}
\SetAlgoLined
\KwResult{Map for area : list of techniques for that area }
 initialization $ T \leftarrow \phi $\;
 \For{$P \in Corpus$}{
  $ A \leftarrow Area (P)$
  $ T \leftarrow \phi $
  $ MSet \leftarrow MethodPapersCitedBy (P)$
  \For {$M \in MSet $}{
  	$T \leftarrow T \cup Technique(M) $
    }
  $T (A) \leftarrow T(A) \cup T $
 }
 \caption{Mapping Areas to List of Techniques}
\end{algorithm}
In Algorithm 1, \textit{Area(P)} is a function that returns the area of a paper P and similarly \textit{Technique(M)} is a function that returns the techniques introduced or improved upon by method paper M. Further, \textit{MethodPapersCitedBy(P)} is a function that returns all the method papers cited by paper P in its methodology section. Thus, basically in this algorithm, we pick each paper P from the corpus, find its area and all the techniques of the method papers that it cites in its methodology section and append all these techniques to the list corresponding to the extracted area for this paper.

We also tried a simple variation of this technique by keeping track of the number of times a particular technique features in an area. This way we also get to know which are the most popular techniques for an area. 
\iffalse{}
Another variation that can be possibly tried is using the ranking information for the top $K$ techniques for a method paper while appending to the set of techniques for an area. Basically, the intuition behind this is higher ranked techniques amongst the top $K$ for method paper M, should be scored higher while appending to the techniques of an area. 
\fi

An example of the kind of mapping that we expect is: machine translation $\rightarrow$ word alignment, gale church algorithm, bleu score, Moses toolkit etc. A detailed set of example entries that we obtained through our experiments are presented in the corresponding results section.

\section{Evaluation Results}
In this paper, we have presented a variety of experiments that require a careful analysis of the results to evaluate the efficacy of the algorithms. The first four stages are crucial to the construction of the information extraction system and thus a critical evaluation is presented. Construction of the mapping table is an outcome of the first four stages and examples have been presented to depict the nature of the mapping table. %The portion on temporal analysis is also depicted as a set of examples giving insights into the nature of the variation of techniques for an area with time. In the subsequent sections we will mainly evaluate the four stages involved in the construction of the information extraction system. 
%In the first step, we have a ranked list of the areas. In this case, we judge how many areas in the top $K$ areas are relevant. By relevancy, in this context, we mean that they are genuine areas of computational linguistics as judged by a domain expert. In the next part, each paper is assigned an area based on the algorithm described previously. In this stage we must find the accuracy of correct assignments. In the third sub-task, we find the set of all techniques used in the computational linguistics domain. In this case, again we have to judge the relevance of the techniques returned in the top K techniques. Here again, relevancy is determined by whether a technique is indeed a genuine technique in computational linguistics as judged by a domain expert. In the last part, for all the method papers, we extracted the specific techniques that they are used for. The accuracy of our algorithm is tested to determine whether the extracted techniques for a given method paper is accurate or not.  For the sake of evaluation, we return the highest scoring technique for each method paper. Each of these results are analyzed in detail in the following sections.

\subsection{Evaluation of the ranked list of areas} 
First, we present the relative performances of the three schemes (in terms of precision) for creation of the ranked list, in Table~\ref{tab0}. Since Scheme 3 gave the highest precision (assignments were judged by an expert) on the list of top-30 areas, we used this scheme for the creation of the ranked list in the subsequent stages.\newline
\begin{table}[h]
\begin{center}
\begin{tabular}{ |c|c| }
  \hline
  Scheme ID&{Precision (\%)} \\
  \hline
  1 & 57 \\
  2 & 73 \\
  3 & 80 \\
  \hline
\end{tabular}
\end{center}
\caption{The precision values for the top 30 areas extracted by various schemes.}
\label{tab0}
\end{table}
\newline
In this part of the evaluation, we have a ranked list of potential areas in the computational linguistics domain extracted from the ACL corpora. We use standard precision-recall measures for the purpose of evaluation. However, note that, there is no easy way to measure the recall of the algorithm on the entire ACL corpora since all the possible areas in the corpus is not known a-priori. This is because the only way to construct a definitive list of all possible areas is to manually identify them from every individual paper in the corpus. However, it is relatively easy to measure the recall of our algorithm on a smaller subset. Therefore, we selected a random set of 200 research papers and manually identified each of their areas. 23 distinct areas were found from this smaller corpus by one domain expert. The areas found out manually were matched to the closest area from the pool of areas. This was done so that we can verify the set of areas returned by the algorithm without ambiguity. This set of areas is the ground-truth for our algorithm. Once this smaller corpus of 200 papers was created, we ran the $k$-gram discounting algorithm with thresholds (Scheme 3) to arrive at a set of areas. Among the 23 cases, 20 of the areas were successfully identified. Thus, the recall of our algorithm is 87 \%. 

Measurement of the precision of our algorithm was relatively straightforward. Measurement of precision can be done by manually judging what fraction of the top $K$ areas are indeed genuine computational linguistics areas as judged by domain experts.  The annotation was done by a domain expert in the following way:  1 if it is a true positive and 0 if it is a false positive. Table~\ref{tab1} notes the values of precision obtained for $K= 25,30,50,75$ and $100$ top application areas. As we can see, as the value of $K$ increases the precision falls which is a witness to the fact that actual areas are ranked higher by our ranking methodology. 
\begin{table}[h]
\begin{center}
\begin{tabular}{ |c|c| }
  \hline
  K&{Precision (\%)} \\
  \hline
  25 & 84  \\
  30 & 80  \\
  50 & 72  \\
  75 & 51  \\
  100 & 43  \\
  \hline
\end{tabular}
\end{center}
\caption{The precision values using Scheme 3 for $K$ = 25, 50, 75 and 100 for extraction of the list of application areas.}
\label{tab1}
\end{table}
\newline
We also asked another domain expert to annotate the first 30 results independent of the first judge. Inter-annotator agreement (Cohen's kappa coefficient) was calculated and the value of $\kappa$ came out to be \textsc{0.79}. The matrix with the agreement/disagreement count between the experts is presented in Table ~\ref{tabar}.
\iffalse $\kappa = \frac{p_{o} - p_{e}}{1- p_{e}}$ \newline
where $p_{o}$ is the relative observed agreement among experts and
$p_{e}$ is the hypothetical probability of chance agreement.\fi 
\begin{table}
\begin{tabular}{l|l|c|c|c|}
\multicolumn{2}{c}{}&\multicolumn{2}{c}{Domain Expert 2}&\multicolumn{1}{c}{}\\
\cline{3-5}
\multicolumn{2}{c|}{}&Yes&No&\multicolumn{1}{c|}{Total}\\
\cline{2-5}
\multirow{2}{*}{Domain Expert 1}& Yes & $23$ & $1$ & $24$\\
\cline{2-5}
& No & $1$ & $5$ & $6$\\
\cline{2-5}
\multicolumn{1}{c|}{} & \multicolumn{1}{c|}{Total} & \multicolumn{1}{|c|}{$24$} & \multicolumn{1}{|c|}{$6$} & \multicolumn{1}{|c|}{$30$}\\
\cline{2-5}
\end{tabular}
\caption{The matrix of agreement and disagreement between two domain experts for annotation of area list.}
\label{tabar}
\end{table}
\subsection{Evaluating the extraction of areas from \\individual papers}
Once we have a probable area for all the papers in the ACL corpus we need to validate the accuracy of our assignment algorithm. Through an online survey, we have conducted the validation of these assignments by a team of domain experts. A set of 120 papers was validated by these domain experts. The accuracy of our method is calculated as \(\frac{x}{120}\) where $x$ is the number of application area assignments marked as correct by the judges. Out of the 120 papers to be evaluated, the judges identified 88 evaluations as correct and hence the accuracy of our method is \textbf{73.3\%}.
\newline
\subsection{Evaluating the list of techniques used in the corpus} 
Evaluating this step is very similar to the evaluation of the pool of application areas generated by our algorithm. In this case, recall calculation is difficult if we work with the top $K$ techniques for each method paper. To simplify the task, we proceed to calculate recall for only the highest ranked technique for each method paper. The experiment is setup in the following way -- for a randomly selected set of 30 papers we aggregate all their citation contexts from the method sections of the citing papers. The smaller corpus of 30 papers introduced or improved upon 26 distinct techniques, is judged by one domain expert. Then we ran the noun phrase extraction algorithm to obtain the list of techniques. 21 of these 26 techniques could be extracted and thus, the recall of our algorithm is around 81\%. 

Precision calculation is %%%again, fairly simple and can be 
done by finding out what percentage of the top $K$ techniques are genuine techniques of the computational linguistics domain, as judged by domain experts. Thus, through manual annotation by a domain expert we found the precision at top $K$ for the technique list at various values of $K$. Table~\ref{tab2} shows the precision obtained for the technique extraction algorithm for various values of $K$. As the value of $K$ increases the precision falls, which is, once again, a witness to the fact that actual areas are ranked higher by our ranking methodology. 

\begin{table}
\begin{center}
\begin{tabular}{ |c|c| }
  \hline
  K&{Precision (\%)} \\
  \hline
  25 & 80 \\
  50 & 64 \\
  75 & 48 \\
  100 & 41 \\
  \hline
\end{tabular}
\end{center}
\caption{The precision values for $K = 25, 50, 75$ and $100$ for extraction of the list of techniques.}
\label{tab2}
\end{table}

Here again we asked another domain expert to annotate the results independent of the first judge. We also calculated the inter-annotator agreement (Cohen's kappa coefficient) for the top 25 techniques and $\kappa$ came out to be \textsc{0.65}. The matrix of agreement/disagreement counts is presented in Table~\ref{tabtech}.
\begin{table}
\begin{tabular}{l|l|c|c|c|}
\multicolumn{2}{c}{}&\multicolumn{2}{c}{Domain Expert 2}&\multicolumn{1}{c}{}\\
\cline{3-5}
\multicolumn{2}{c|}{}&Yes&No&\multicolumn{1}{c|}{Total}\\
\cline{2-5}
\multirow{2}{*}{Domain Expert 1}& Yes & $18$ & $2$ & $20$\\
\cline{2-5}
& No & $1$ & $4$ & $5$\\
\cline{2-5}
\multicolumn{1}{c|}{} & \multicolumn{1}{|c|}{Total} & \multicolumn{1}{|c|}{$19$} & \multicolumn{1}{|c|}{$6$} & \multicolumn{1}{|c|}{$25$}\\
\cline{2-5}
\end{tabular}
\caption{ The matrix of agreement and disagreement between two domain experts for annotation of technique list.}
\label{tabtech}
\end{table}

\subsection{Evaluating the extraction of techniques\\ from a method paper}
Once we have the repertoire of tasks for the entire corpus by using the dot-product based ranking we can identify the specific techniques a method paper is used for. We now need to evaluate the accuracy of this assignment of techniques for a method paper. For simplicity of the evaluation, we report only the top ranked technique for a method paper. For the evaluation, a sample of 60 papers were selected randomly and the judgments of a set of domain experts collected through an online survey were used to test the accuracy of our method. Thus, each domain expert evaluated the technique assigned for the papers submitted to him/her and reported how many of these were correct. The accuracy of our algorithm is then calculated as \( \frac{x}{60} \), where $x$ is the number of correct assignments of technique to the method papers. In this evaluation, our algorithm had 36 correct assignments and thus the accuracy of our method is \textbf{60\%}.

\subsection{Mapping from areas to list of techniques}
Earlier, we discussed the procedure for the construction of the mapping from areas to the list of techniques. In this section we present some of the entries that we obtained from the mapping table for some of the higher ranked areas of computational linguistics. 
As we see from the examples, the techniques extracted consists of sub-tasks, tools and data-sets popularly used in an area. Further, the techniques are quite detailed and describe many techniques which are very specific to the areas - for example, bleu score and gale church algorithm : machine translation, nivres arc eager : dependency parsing and spin model : opinion mining. Also, it is interesting to observe that the extracted techniques span a wide range of time- for example: Collins parser, Berkeley parser, Charniak parser, Stanford parser, MST parser and Malt parser : Dependency Parsing were introduced to the computational linguistics (CL) community at substantially different time periods.

\begin{table*}[!thb]
 \resizebox{0.8\textwidth}{!}{\begin{minipage}{\textwidth}
 \begin{tabular}{|c|c| }
  \hline
  \textbf{Area} & \textbf{Techniques}\\
  \hline
  Machine Translation & {\parbox[t]{17cm}{Bleu score, rate training, IBM model, word alignment, Moses toolkit, inversion transduction grammar,bootstrap resampling, translation model, penn treebank, translation quality,language model, gale church algorithm}}  \\
  \hline
 Dependency Parsing & {\parbox[t]{17cm}{penn treebank, malt parser, berkeley parser, mst parser, charniak parser, collins parser, maximum entropy, nivres arc eager, stanford parser, perceptron algorithm }}\\
 \hline
 Multi-document Summarization & {\parbox[t]{17cm}{topic signatures, information extraction, page rank, klsum summarization system, mead summarizer, word sense disambiguation, lexical chains, inverse sentence frequency}}\\
 \hline
 Word Sense Disambiguation & {\parbox[t]{17cm}{coarse senses,semcor corpus, senseval competitions, semantic similarity, micro context, maximum entropy, mutual information}}\\
 \hline
 Social Media & {\parbox[t]{17cm}{natural language processing, entity recognition,support/oppose camps, sentiment analysis, automatic translation, mutual information, stanford parser, website twitter}}\\
 \hline
 Sense Induction & {\parbox[t]{17cm}{word sense disambiguation, semeval word sense induction,chinese whispers, recursive spectral clustering, topic models, graded sense annotation,ontonotes project}}\\
 \hline
 Opinion Mining & {\parbox[t]{17cm}{sentiment analysis, mutual information, spin model, subjectivity lexicon, semantic role analysis, multiclass clasifier, coreference resolution, latent dirichlet}}\\
 \hline
 Chinese Word Segmentation & {\parbox[t]{17cm}{entity recognition, conditional random fields, segmentation bakeoff, stanford chinese word segmenter, perceptron algorithm , discourse segmentation, crf model}}\\
  \hline
\end{tabular}
\end{minipage}}
\caption{Mapping from areas to list of techniques: Example application areas and corresponding techniques from AAN dataset.}
\label{tab1}
\end{table*}

\section{Use Case: Temporal analysis} 
\label{use-case}
In this section, we present three use cases. Each use case is a temporal study of areas and techniques. We analyze evolution of 
application areas and corresponding techniques over a given time-period. 
In the first use case, we study the evolution of areas in the computational linguistics field. Next use case deals with evolution of 
techniques for a given area. Lastly, we study the evolution of major areas in the top conferences. 
\begin{center}
\begin{figure}[h]
\includegraphics[width=0.47\textwidth, height=5cm]{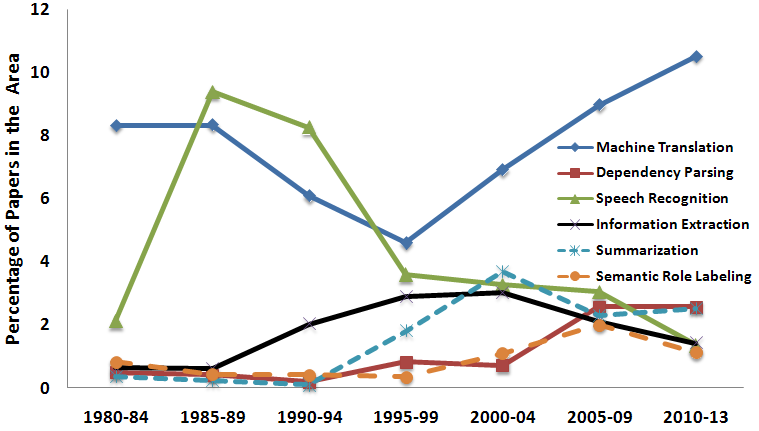}
\caption{Evolution of different application areas over time in terms of fraction of publications.}
\label{graph}
\end{figure}
\end{center}

\subsection{Evolution of areas}
From the list of popular areas (based on the total number of papers published in an area) in \textsc{aan}, we present six representative areas, namely, ``machine translation'', ``dependency parsing'', ``speech recognition'', ``information extraction'', ``summarization'' and ``semantic role labeling'', and study their popularity (percentage of papers in that area for that time period out of total papers published in that time period) from 1980-2013 in 5-year windows. Figure~\ref{graph} demonstrates the temporal variations for these areas and how they evolve with time. 

\noindent{\bf Observations:} While areas like ``machine translation'' and ``dependency parsing'' are on the rise, ``information extraction'' and ``semantic role labeling'' are on a decline. A further interesting observation is that till 1994 the ACL community had a lot of interest in ``speech recognition'' which then saw a sharp decline possibly because of the fact that the speech community slowly separated out.

\subsection{Evolution of techniques in areas}
In the second use case, we study evolution of techniques for a given area. For this analysis, we divide the time-line 
into fixed buckets of $w$ years. Next, for each bucket, we extract popular techniques (based on the number of times any paper has cited that technique) using our proposed system. Interestingly, we found multiple trends 
and figures in this study. Table~\ref{tab:technique-in-area} presents the popular techniques for five example areas. Some of the interesting trends
from the table~\ref{tab:technique-in-area} are listed below: 
\begin{itemize}
\item \textbf{Dependency Parsing:} New techniques (Malt parser, minimum spanning tree (MST) parser, etc.) came into existence in 2005 -- 2009. In the next year bucket, these parsers overcome popularity of previous parsers such as Collin's parser, Berkeley parser and are almost at par with Charniak parser. In addition, we observe that the Penn treebank is extensively used for dependency parsing across almost all time periods.
\item \textbf{Machine Translation:} We found that word alignment and inversion transduction grammar are popular techniques for machine translation across all time periods. Also, Bleu score has been a popular technique since its introduction in 2000 -- 2004. Similarly, Moses toolkit and IBM model are both popular techniques across most time periods.
\item \textbf{Sentiment Analysis:} In this area, mutual information and word sense disambiguation are popular techniques for most of the time periods. Latent dirichlet allocation (introduced in 2003) found important use in sentiment analysis in 2005 -- 2009.  Also the spin model got popularity in 2005 -- 2009.
\item \textbf{Cross Lingual Textual Entailment:} Distributional similarity and mutual information are important techniques and are popular in multiple time periods. Verb ocean gets popular in 2005 -- 2009 and 2009 -- 2013. It is also very interesting to note that machine translation is actually an important tool for this area and is very popular in 2005 -- 2009. However, in 2010 -- 2013 its popularity goes down. A probable explanation for this could be the introduction of techniques which perform cross-lingual textual entailment without machine translation - for example, \textit{FBK: Cross-Lingual Textual Entailment Without Translation}~\cite{mehdad2012fbk}.
\item \textbf{Grammatical Error Correction}: Techniques to address out-of-Vocabulary (OOV) words have become important in recent times. Over the years, Collins parser got replaced by Charniak parser and finally by Berkeley parser. Penn treebank is an important dataset for this area.
\end{itemize}

\begin{table*}

 \resizebox{0.75\textwidth}{!}{\begin{minipage}{\textwidth}
\begin{tabular}{ |c|c|c|c|c|c|}
  \hline
  \multirow{2}{*}{Area} & \multicolumn{5}{|c|}{Techniques} \\
   \cline{2-6}
  & 1990-1994&1995-1999&2000-2004&2005-2009&2010-2013\\
  \hline
{\parbox[t]{2cm}{Dependency Parsing}} & {\parbox[t]{3.5cm}{dependency unification grammar, kasper algorithm, left corner parser, inheritence systems, eurotra project}} & {\parbox[t]{3.5cm}{penn treebank, probabilistic context free grammar, tree substitution grammar, conditional random fields, dependency links,collins parser, dependency unification grammar, berkeley parser, maruyamas constraint dependency grammar}} & {\parbox[t]{3.5cm}{penn treebank, collins parser, berkeley parser, charniak parser, maximum entropy, negra corpus}} & {\parbox[t]{3.5cm}{penn treebank, charniak parser, malt parser, mst parser, berkeley parser, stanford parser, ccg parser, nivres arc eager,maximum entropy, perceptron algorithm}} & {\parbox[t]{3.5cm}{penn treebank, malt parser,mst parser, berkeley parser, charniak parser, stanford parser,perceptron algorithm, nivres arc eager}}\\
\hline
 {\parbox[t]{2cm}{Machine Translation}} & {\parbox[t]{3.5cm}{parse parse match approaches, early type deduction, bottom up head driven algorithm, bilingual signs}} & {\parbox[t]{3.5cm}{ibm model, inversion transduction grammar, word alignment, sentence alignment, parse parse match approaches,moses toolkit, brill tagger}} &{\parbox[t]{3.5cm}{word alignment, bleu score, inversion transduction grammar, parse parse match approaches}} & {\parbox[t]{3.5cm}{rate training, ibm model, bleu score, word alignment, inversion transduction grammar,moses toolkit}} & {\parbox[t]{3.5cm}{bleu score, rate training, moses toolkit, word alignment,bootstrap resampling, ibm model, language model}}\\
\hline
{\parbox[t]{2cm}{Sentiment Analysis}} & {\parbox[t]{3.5cm}{early type deduction mechanisms, unification grammars,sentence plan language, mutual information, taxonomy files}} & {\parbox[t]{3.5cm}{levenshtein distance, discourse structure}} & {\parbox[t]{3.5cm}{mutual information, information extraction, penn treebank, distributional similarity,statistical parser, word sense disambiguation}} & {\parbox[t]{3.5cm}{mutual information, word sense disambiguation, subjectivity lexicon, latent dirichlet , spin model, relation extraction, movie reviews}} & {\parbox[t]{3.5cm}{mutual information, word sense disambiguation, subjectivity lexicon, latent dirichlet, polarity lexicons, movie reviews, support/oppose camps}} \\
\hline
{\parbox[t]{2cm}{Cross-lingual Textual Entailment}} & {\parbox[t]{3.5cm}{ordinary dictionary, text generation, dependency unification grammar, machine translation, mutual information}} & {\parbox[t]{3.5cm}{discourse structure, encode tfs, temporal information,english texts, kappa coefficient, cue phrases,brill tagger}} & {\parbox[t]{3.5cm}{mutual information, manual annotation, distributional similarity,heuristic approaches, lexical chains,log linear model}} & {\parbox[t]{3.5cm}{word sense disambiguation, machine translation,textual entailment challenge,distributional similarity, mutual information,verb ocean}} & {\parbox[t]{3.5cm}{semantic textual similarity, verb ocean,moses toolkit, machine translation}}\\
\hline{\parbox[t]{2cm}{Grammatical Error Correction}} & {\parbox[t]{3.5cm}{probabilistic context free grammars, parseval metric, brill pos tagger}} & {\parbox[t]{3.5cm}{penn treebank, prepositional phrase attachment,collins parser}} & {\parbox[t]{3.5cm}{penn treebank,brill tagger, fntbl toolkit, charniak parser, kappa statistics}} & {\parbox[t]{3.5cm}{penn treebank,word sense disambiguation, charniak parser, oov words}} & {\parbox[t]{3.5cm}{english corpus, clc fce dataset, oov words, berkeley parser, charniak parser}}\\\hline

\end{tabular}
\end{minipage}}
\caption{A few examples of areas and their top techniques for different time periods.}
\label{tab:technique-in-area}
\end{table*}

\begin{figure*}
\centering
 \resizebox{0.9\textwidth}{!}{\begin{minipage}{\textwidth}
\begin{tabular}{@{}c@{}c@{}c@{}c@{}c@{}c@{}c@{}}
  && &AAN& && \\
\includegraphics[width=.23\hsize]{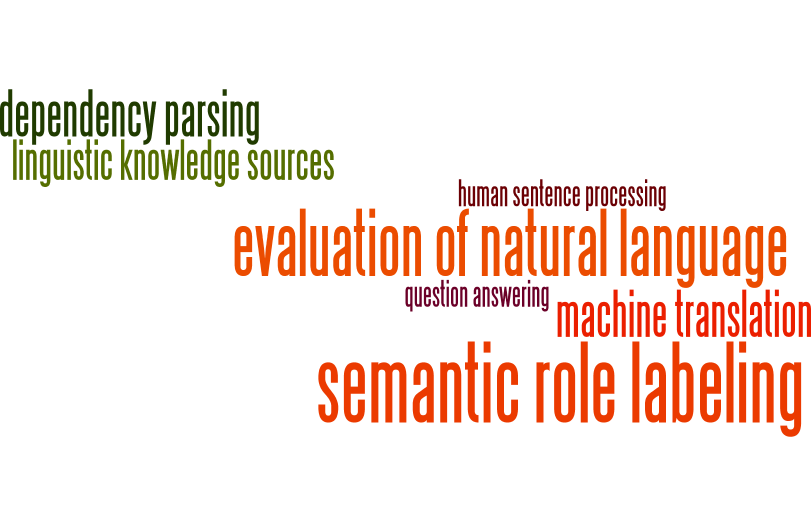} & \;\;\;&
 \includegraphics[width=.23\hsize]{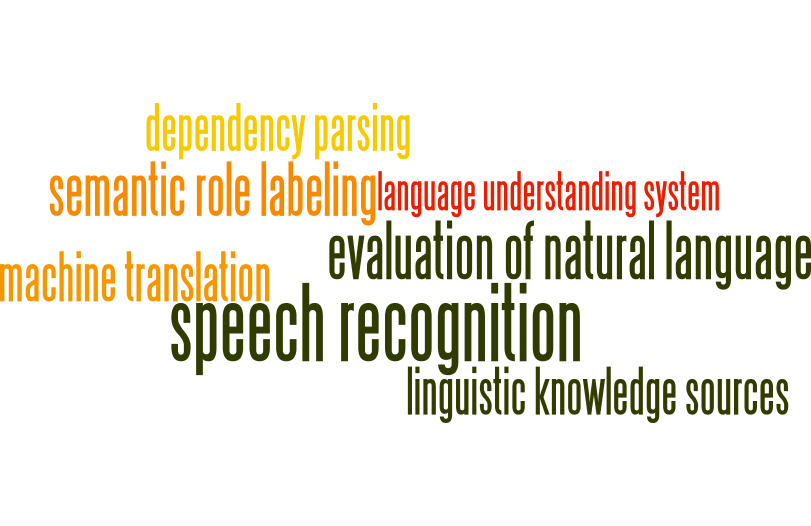} &\hspace{-20cm} &
 \includegraphics[width=.23\hsize]{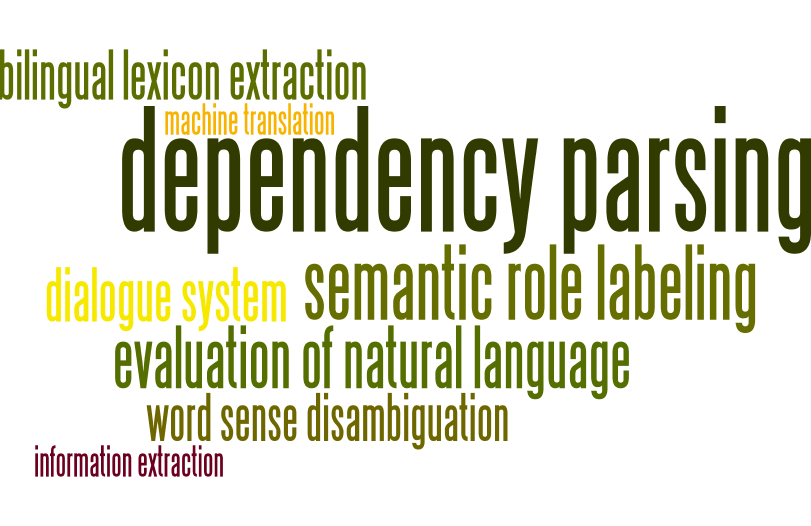} &  \;\;\;&
 \includegraphics[width=.23\hsize]{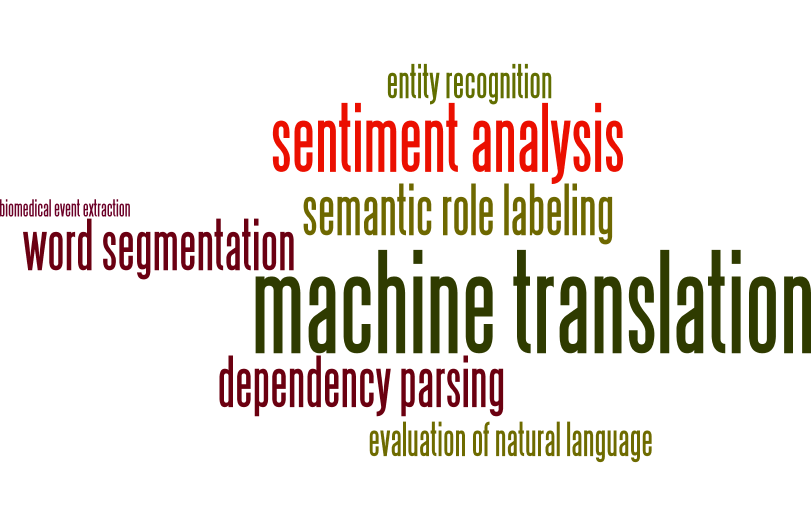} \\
  1975-1984 &&1985-1994 && 1995-2004 &&2005-2013 \\
  && && && \\
 
   && &ACL& && \\
  \includegraphics[width=.23\hsize]{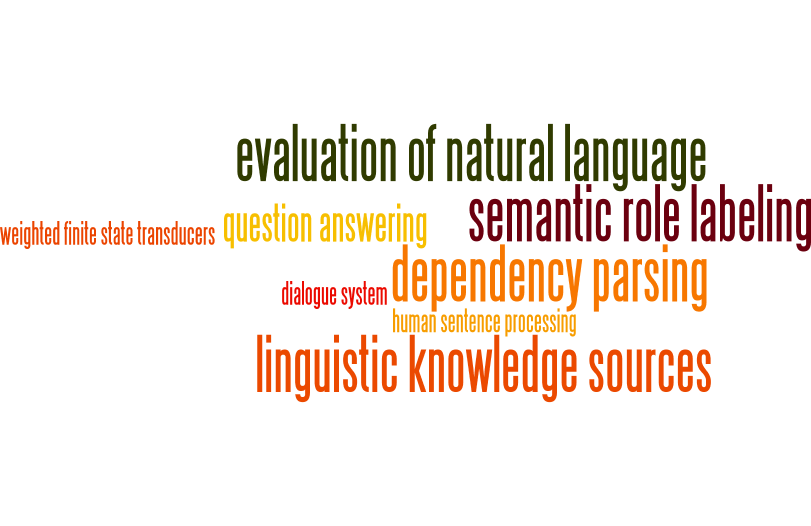} &&
  \includegraphics[width=.23\hsize]{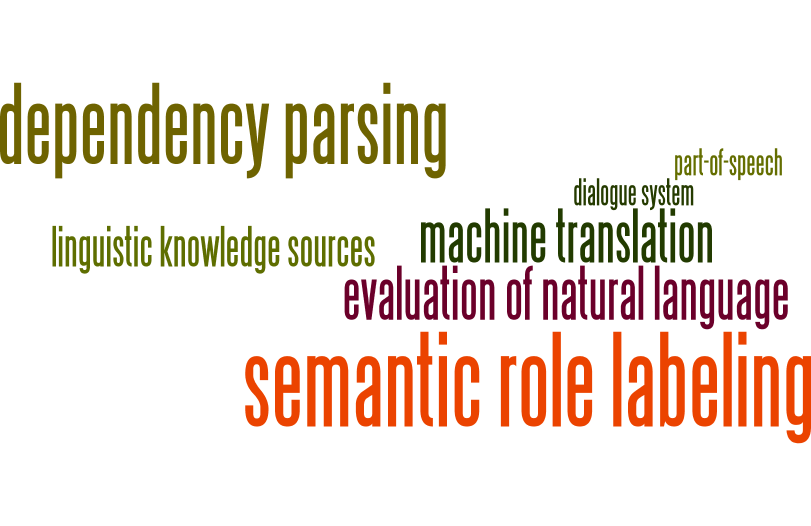} &&
  \includegraphics[width=.23\hsize]{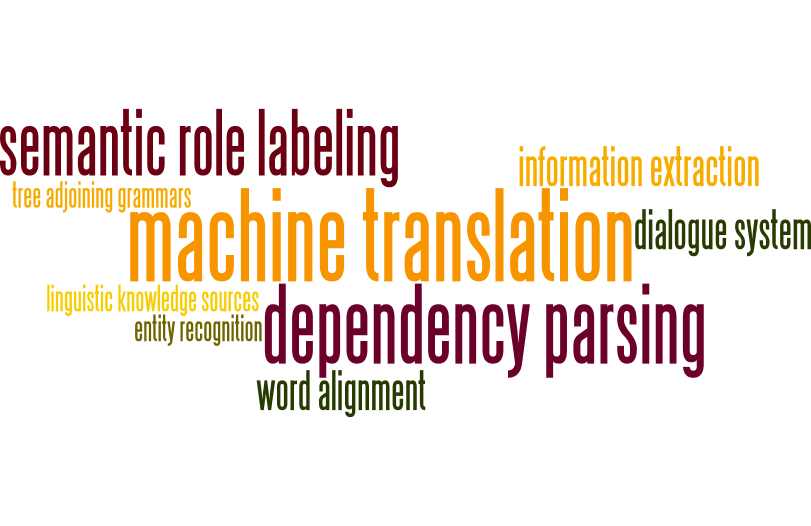} &&
  \includegraphics[width=.23\hsize]{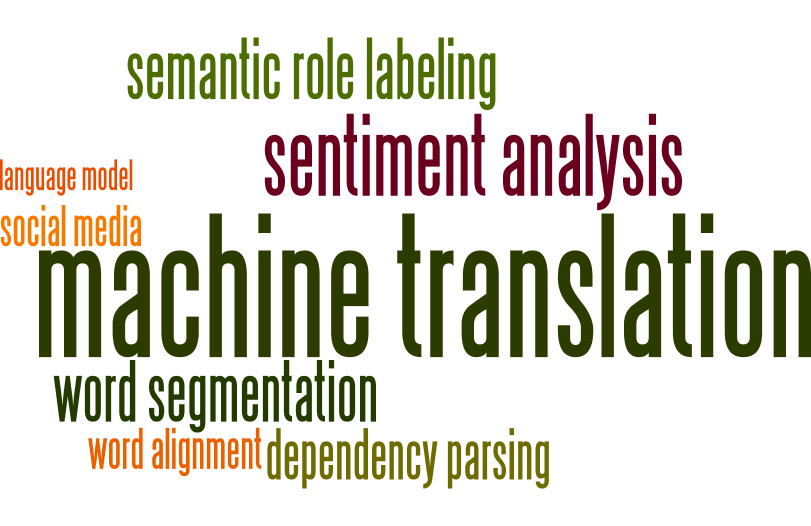} \\
  1975-1984 &&1985-1994 && 1995-2004 && 2005-2013 \\
  && && && \\

    && &COLING& &&  \\
  \includegraphics[width=.23\hsize]{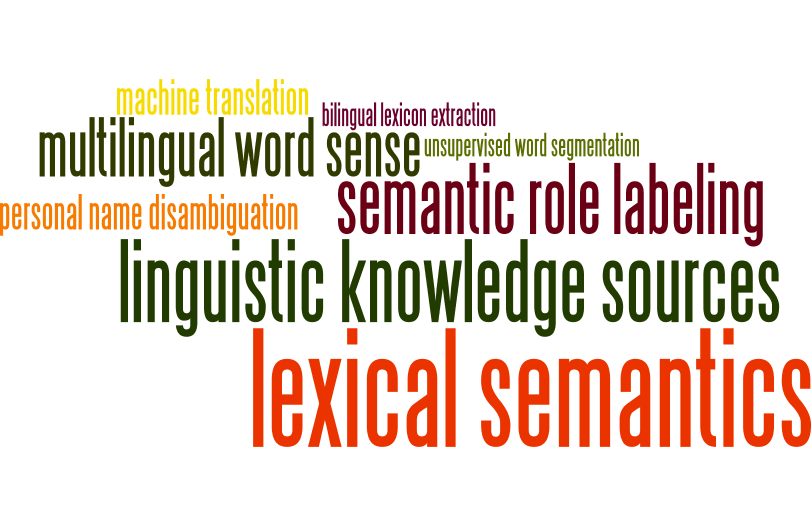} &&
  \includegraphics[width=.23\hsize]{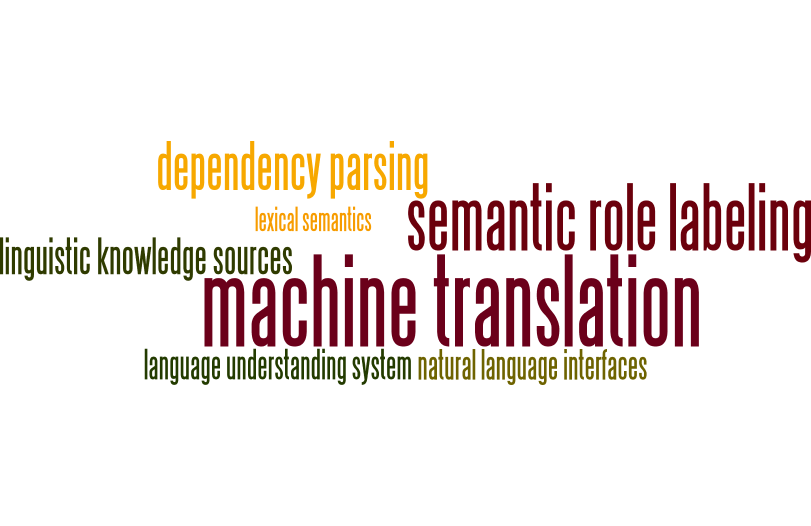} &&
   \includegraphics[width=.23\hsize]{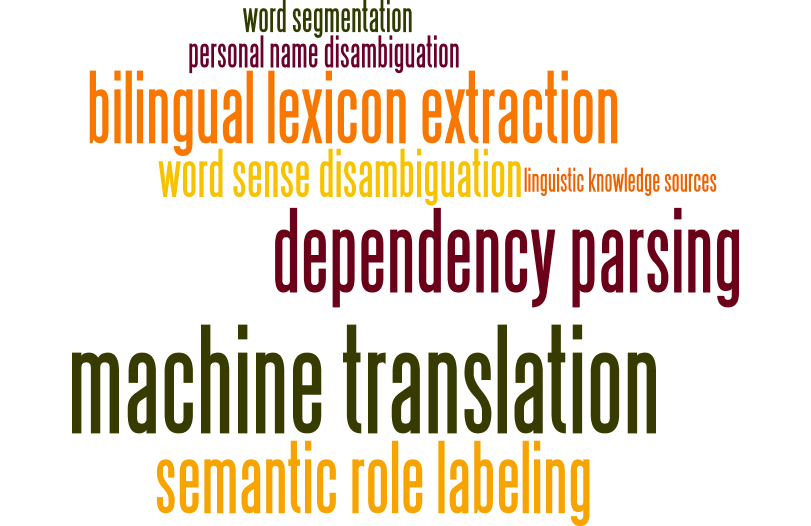} &&
  \includegraphics[width=.23\hsize]{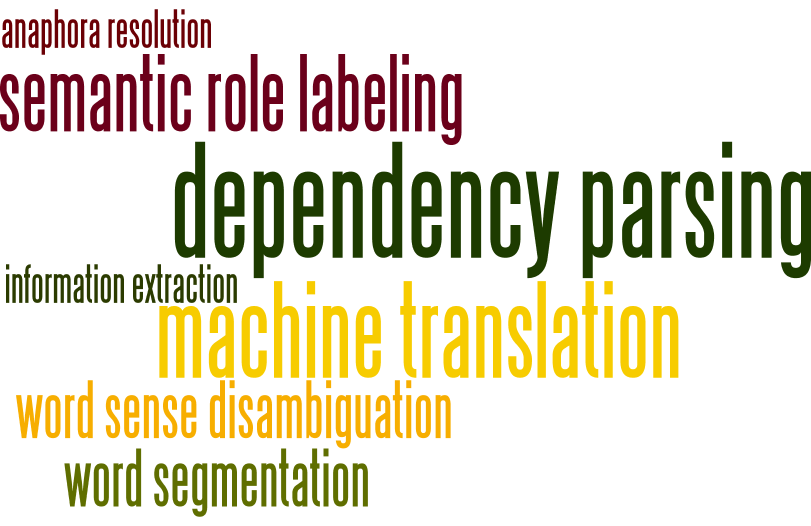} \\
  1965-1974 && 1975-1984 && 1985-1994 && 1995-2004 \\
  
\end{tabular}
\end{minipage}}
\label{fig:word_cloud}
\caption{Phrase-Clouds representing the proportion of papers for an area across time periods for \textsc{AAN} dataset, \textsc{ACL} and \textsc{COLING}.}  
\label{fig:word_cloud}
\end{figure*}

%{\color{red} SOHAM: can you please update the plots since they are not very visible}

\subsection{Evolution of major areas in top conferences}
We shortlisted two top conferences in the computational linguistics domain namely, Annual Meeting of the Association of Computational Linguistics (ACL)
and the International Conference on Computational Linguistics (COLING). We study 40 years of conference history. For each conference, 40 year time-period
 is divided into four 10-year buckets. Next, for each conference, we extract top ten most popular areas (based on citation counts) for each 10-year bucket.
 Figure \ref{fig:word_cloud} presents phrase clouds representing evolution of areas in these two conferences in comparison to the full AAN dataset itself. Some of the interesting observations from this analysis are:
\begin{itemize}
\item \textbf{Full AAN dataset:} Here, we observe that while in the earlier decades, areas such as ``semantic role labeling'', ``evaluation of natural language'' and ``speech recognition'' were dominant, they fade away in the recent decades. On the other hand, areas such as ``machine translation'' and ``dependency parsing'', which were less prevalent in the earlier decades gain significant importance in the recent decades. We also see ``sentiment analysis'' as one of the major areas in the last decade.
\item \textbf{ACL:} In the earlier decades of this conference, one finds that the community is interested in areas like ``linguistic knowledge sources'' and ``semantic role labeling.'' Over the recent decades, however, the community seems to be more interested in areas such as ``machine translation'' and ``dependency parsing''. Interestingly, in the time period 2005 -- 2013, an upcoming area like ``social media'' is found to gain importance.
\item \textbf{COLING:} For this conference, we observe that in the earlier decades, areas like ``lexical semantics'' and ``linguistic knowledge sources'' were of interest to the community. However, in the recent years, areas like ``machine translation'', ``dependency parsing'' and ``bilingual lexicon extraction'' have gained importance. An interesting observation here is that ``semantic role labeling'' has been all through a thrust area for this particular conference. 

\end{itemize}

\section{Conclusion and Future Work}
In this paper, we have proposed a novel information extraction system for scientific articles. The system extracts ranked list of all application areas in the computational linguistics domain. At a more granular level, it also extracts application area for a given paper. In addition, it extracts ranked list of 
all techniques as well as paperwise technique extraction. Finally, we construct a mapping from application areas to all the techniques. We evaluate our system with domain experts and prove that it performs reasonably well on both precision and recall. As a use case, we present an extensive analysis of temporal variation in popularity of the {\em techniques} for a given area. Some of the interesting observation that we make here are that the areas like ``machine translation'' and ``dependency parsing'' are on the rise of popularity while areas like ``speech recognition'', ``linguistic knowledge sources'' and ``evolution of natural language'' are on the decline.  

In future, we plan to work on constructing a multi-level mapping table that maps application areas to techniques and further techniques 
to a set of parameters. For example, application area \textit{Machine Translation} has \textit{Bleu score}
 as one of its techniques. Bleu score is a algorithm that takes few input parameters. Changing these parameters will change the outcome
 of the score. Example of one such parameter is $n$, which represents the value of $n$ for the $n$-grams. 

We also plan to run the current system on a larger corpus of scientific articles. All our methods can be generalized to domains other than computational 
linguistics. Currently, we plan to extend the dataset to Microsoft academic research dataset (larger dataset) and a biomedical dataset (different domain).
We also plan to study temporal characteristics of techniques for a given application area to observe if future predictions can be made 
for a technique - whether its popularity will increase or decrease in the years come.

\bibliographystyle{abbrv}
\bibliography{sig-alternate-sample} 
\end{document}